\documentclass[11pt]{article}

\usepackage{acl2016}
\usepackage{times}
\usepackage{latexsym}

\usepackage{microtype}
\usepackage{url}
\usepackage{breakurl} 

\usepackage{textcomp}
\usepackage{amsmath}
\usepackage{amssymb}
\usepackage{threeparttable}
\usepackage{comment}
\usepackage[labelsep=newline]{subfig}
\usepackage{soul}
\usepackage{color}
\usepackage[table]{xcolor}
\definecolor{green4}{HTML}{21610B}
\definecolor{green3}{HTML}{31B404}
\definecolor{green2}{HTML}{00FF00}
\definecolor{green1}{HTML}{81F781}
\definecolor{red4}{HTML}{B40404}
\definecolor{red3}{HTML}{FE2E2E}
\definecolor{red2}{HTML}{FA5858}
\definecolor{red1}{HTML}{F78181}

\newcommand{\hlc}[2][yellow]{{\sethlcolor{#1}\hl{#2}}}

\newcommand{\wpr}{\texttt{word2vec$_{\textrm{pre-trained}}$}}
\newcommand{\mbf}[1]{\mathbf{#1}}

\usepackage{pgf,tikz}
\usepackage{mathrsfs}
\usetikzlibrary{arrows,automata}
\usetikzlibrary{decorations.pathreplacing,decorations.pathmorphing}
\usetikzlibrary{decorations.markings}

\tikzset{
  big arrow/.style={
    decoration={markings,mark=at position 1 with {\arrow[scale=1.5,#1]{>}}},
    postaction={decorate},
    shorten >=0.4pt},
  big arrow/.default=black}
\definecolor{qqqqff}{rgb}{0.3333333333333333,0.3333333333333333,0.3333333333333333}
\newcommand{\nods}[2]{
\foreach \x / \y [count=\xi] in {#1/#2,#1+1.75/#2+1.5,#1+3.5/#2+3} {
  \path (\x, -1) coordinate (n1);
  \pgfmathparse{int(mod(\xi,2))}
  \ifnum\pgfmathresult=0{\node (n1) at (\x, -1) {\dots};}
  \else{\node[draw, circle] (n1) at (\x, -1) {};}\fi
  \draw[big arrow] (n1) -- (\y,0);
}}

\aclfinalcopy
\def\aclpaperid{805}

\title{Automatic Text Scoring Using Neural Networks}

\author{Dimitrios Alikaniotis \\
  Department of Theoretical \\
  and Applied Linguistics \\
  University of Cambridge \\
  Cambridge, UK \\
  {\tt da352@cam.ac.uk} \\\And
  Helen Yannakoudakis \\
  The ALTA Institute \\
  Computer Laboratory \\
  University of Cambridge \\
  Cambridge, UK \\  
  {\tt hy260@cl.cam.ac.uk} \\\And
  Marek Rei \\
  The ALTA Institute \\
  Computer Laboratory \\
  University of Cambridge \\
  Cambridge, UK \\
  {\tt mr472@cl.cam.ac.uk} \\}

\date{}
\usepackage{graphicx}

\begin{document}

\maketitle

\begin{abstract}

  Automated Text Scoring (ATS) provides a cost-effective and
  consistent alternative to human marking. However, in order to
  achieve good performance, the predictive features of the system need
  to be manually engineered by human experts. We introduce a model
  that forms word representations by learning the extent to which
  specific words contribute to the text's score.  Using Long-Short
  Term Memory networks to represent the meaning of texts, we
  demonstrate that a fully automated framework is able to achieve
  excellent results over similar approaches. In an attempt to make our
  results more interpretable, and inspired by recent advances in
  visualizing neural networks, we introduce a novel method for
  identifying the regions of the text that the model has found more
  discriminative.

\end{abstract}

\section{Introduction}
\label{sec:intro}

Automated Text Scoring (ATS) refers to the set of statistical and
natural language processing techniques used to automatically score a
text on a marking scale. The advantages of ATS systems have been
established since Project Essay Grade (PEG) \cite{Page1967,Page1968},
one of the earliest systems whose development was largely motivated by
the prospect of reducing labour-intensive marking activities. In
addition to providing a cost-effective and efficient approach to
large-scale grading of (extended) text, such systems ensure a
consistent application of marking criteria, therefore facilitating
equity in scoring.

There is a large body of literature with regards to ATS systems of
text produced by non-native English-language learners \cite[among
others]{Page1968,Attali2006,Rudner2002,elliot:2003,Landauer2003,%
Briscoe2010,DBLP:conf/acl/YannakoudakisBM11,sakaguchi2015effective},
overviews of which can be found in various studies
\cite{williamson:2009,Dikli2006,Shermis2012}. Implicitly or
explicitly, previous work has primarily treated text scoring as a
supervised text classification task, and has utilized a large
selection of techniques, ranging from the use of syntactic parsers,
via vectorial semantics combined with dimensionality reduction, to
generative and discriminative machine learning.

As multiple factors influence the quality of texts, ATS systems
typically exploit a large range of textual features that correspond to
different properties of text, such as grammar, vocabulary, style,
topic relevance, and discourse coherence and cohesion. In addition to
lexical and part-of-speech (POS) $n$grams, linguistically deeper
features such as types of syntactic constructions, grammatical
relations and measures of sentence complexity are among some of the
properties that form an ATS system's internal marking criteria. The
final representation of a text typically consists of a vector of
features that have been manually selected and tuned to predict a score
on a marking scale.

Although current approaches to scoring, such as regression and
ranking, have been shown to achieve performance that is
indistinguishable from that of human examiners, there is substantial
manual effort involved in reaching these results on different domains,
genres, prompts and so forth. Linguistic features intended to capture
the aspects of writing to be assessed are hand-selected and tuned for
specific domains. In order to perform well on different data, separate
models with distinct feature sets are typically tuned.

Prompted by recent advances in deep learning and the ability of such
systems to surpass state-of-the-art models in similar areas
\cite{Tang_2015,1503.00075v3}, we propose the use of recurrent neural
network models for ATS. Multi-layer neural networks are known for
automatically learning useful features from data, with lower layers
learning basic feature detectors and upper levels learning more
high-level abstract features \cite{Lee2009}.  Additionally, recurrent
neural networks are well-suited for modeling the compositionality of
language and have been shown to perform very well on the task of
language modeling \cite{Kombrinka,Chelba2014}. We therefore propose to
apply these network structures to the task of scoring, in order to
both improve the performance of ATS systems and learn the required
feature representations for each dataset automatically, without the
need for manual tuning. More specifically, we focus on predicting a
holistic score for extended-response writing items.\footnote{The task
  is also referred to as Automated Essay Scoring. Throughout this
  paper, we use the terms \textit{text} and \textit{essay} (scoring)
  interchangeably. }

However, automated models are not a panacea, and their deployment
depends largely on the ability to examine their characteristics,
whether they measure what is intended to be measured, and whether
their internal marking criteria can be interpreted in a meaningful and
useful way. The deep architecture of neural network models, however,
makes it rather difficult to identify and extract those properties of
text that the network has identified as discriminative. Therefore, we
also describe a preliminary method for visualizing the information the
model is exploiting when assigning a specific score to an input text.

\section{Related Work}
\label{sec:related_work}

In this section, we describe a number of the more influential and/or
recent approaches in automated text scoring of non-native
English-learner writing.

Project Essay Grade \cite{Page1967,Page1968,Page2003} is one of the
earliest automated scoring systems, predicting a score using linear
regression over vectors of textual features considered to be proxies
of writing quality. Intelligent Essay Assessor \cite{Landauer2003}
uses Latent Semantic Analysis to compute the semantic similarity
between texts at specific grade points and a test text, which is
assigned a score based on the ones in the training set to which it is
most similar. \newcite{Lonsdale2003} use the Link Grammar parser
\cite{Sleator1995} to analyse and score texts based on the average
sentence-level scores calculated from the parser's cost vector.

The Bayesian Essay Test Scoring sYstem \cite{Rudner2002} investigates
multinomial and Bernoulli Naive Bayes models to classify texts based
on shallow content and style features. e-Rater \cite{Attali2006},
developed by the Educational Testing Service, was one of the first
systems to be deployed for operational scoring in high-stakes
assessments. The model uses a number of different features, including
aspects of grammar, vocabulary and style (among others), whose weights
are fitted to a marking scheme by regression.

\newcite{Chen2010} use a voting algorithm and address text scoring
within a weakly supervised bag-of-words framework.
\newcite{DBLP:conf/acl/YannakoudakisBM11} extract deep linguistic
features and employ a discriminative learning-to-rank model that
outperforms regression.

Recently, \newcite{mcnamara2015hierarchical} used a hierachical
classification approach to scoring, utilizing linguistic, semantic and
rhetorical features, among others. \newcite{farra2015scoring} utilize
variants of logistic and linear regression and develop models that
score persuasive essays based on features extracted from opinion
expressions and topical elements.

There have also been attempts to incorporate more diverse features to
text scoring models. \newcite{klebanov2013word} demonstrate that essay
scoring performance is improved by adding to the model information
about percentages of highly associated, mildly associated and
dis-associated pairs of words that co-exist in a given text.
\newcite{somasundaran2014lexical} exploit lexical chains and their
interaction with discourse elements for evaluating the quality of
persuasive essays with respect to discourse coherence.
\newcite{crossley2015pssst} identify student attributes, such as
standardized test scores, as predictive of writing success and use
them in conjunction with textual features to develop essay scoring
models.

In 2012, Kaggle,\footnote{\texttt{http://www.kaggle.com/c/asap-aes/}}
sponsored by the Hewlett Foundation, hosted the Automated Student
Assessment Prize (ASAP) contest, aiming to demonstrate the
capabilities of automated text scoring systems
\cite{shermis2015contrasting}. The dataset released consists of around
twenty thousand texts ($60\%$ of which are marked), produced by
middle-school English-speaking students, which we use as part of our
experiments to develop our models.

\section{Models}
\label{sec:models}

\subsection{C\&W Embeddings}
\label{sec:cw-embeddings}

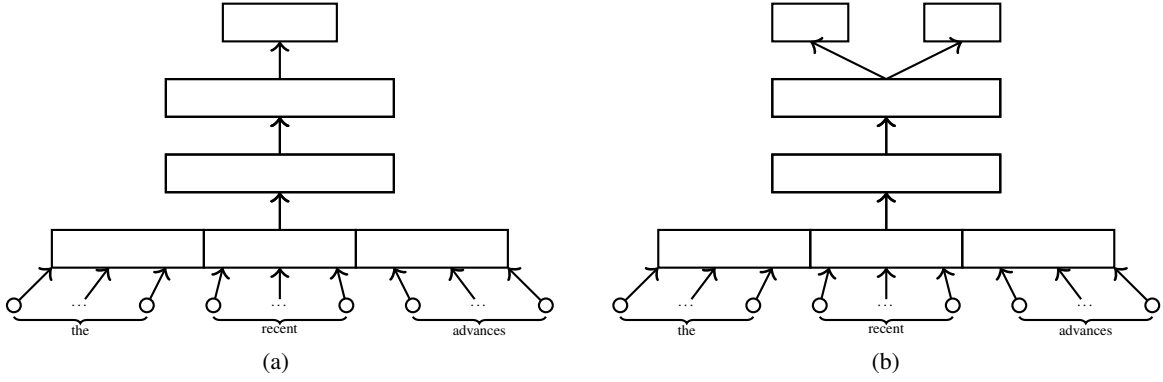
\begin{figure*}[t]
  \centering  
  \subfloat[][]{
    \begin{tikzpicture}[thick,scale=0.5, every node/.style={scale=0.5}]
      \foreach \x in {0,4,...,9} \draw (\x, 0) rectangle (\x+4, 1);
      \foreach \x in {1,3} \foreach \y in {2,4} {
        \draw[big arrow] (6, \x) -- (6, \x+1);
        \draw (3, \y) rectangle (9, \y+1);
      }
      \draw[big arrow] (6, 5) -- (6, 6);
      \draw (4.5, 6) rectangle (7.5, 7);
      \foreach \xj / \yj in {-1/0,4.25/4.5,9.5/9} \nods{\xj}{\yj};
      \foreach \x/\xtext in {-1/the,4.25/recent,9.5/advances}
      \draw[decorate,thick,decoration={brace,mirror}] (\x,-1.3) -- node[below=2pt] {\xtext} (\x+3.5,-1.3);
    \end{tikzpicture}\label{fig:cw}}\quad\quad
  \subfloat[][]{\begin{tikzpicture}[thick,scale=0.5, every node/.style={scale=0.5}]
  \foreach \x in {0,4,...,9} \draw (\x, 0) rectangle (\x+4, 1);
  \foreach \x in {1,3} \foreach \y in {2,4} {
    \draw[big arrow] (6, \x) -- (6, \x+1);
    \draw (3, \y) rectangle (9, \y+1);
  }
  \foreach \x in {3,7} {
    \draw (\x, 6) rectangle (\x+2, 7);
    \draw[big arrow] (6, 5) -- (\x+1, 6);
  }
  \foreach \xj / \yj in {-1/0,4.25/4.5,9.5/9} \nods{\xj}{\yj};
  \foreach \x/\xtext in {-1/the,4.25/recent,9.5/advances}
  \draw[decorate,thick,decoration={brace,mirror}] (\x,-1.3) -- node[below=2pt] {\xtext} (\x+3.5,-1.3);
\end{tikzpicture}\label{fig:sswe}}\quad\quad
  \caption{Architecture of the original C\&W model (left) and of our
    extended version (right).}
  \label{fig:architectures}
\end{figure*}

\newcite{Collobert:2008kg} and \newcite{1103.0398v1} introduce a
neural network architecture (Fig.~\ref{fig:cw}) that learns a
distributed representation for each word $w$ in a corpus based on its
local context. Concretely, suppose we want to learn a representation
for some target word $w_t$ found in an $n$-sized sequence of words
$\mathcal{S} = \left ( w_1, \dots, w_t, \dots, w_n \right )$ based on
the other words which exist in the same sequence $\left ( \forall w_i
\in \mathcal{S}\,|\, w_i \neq w_t \right )$. In order to derive this
representation, the model learns to discriminate between $\mathcal{S}$
and some `noisy' counterpart $\mathcal{S}^\prime$ in which the target
word $w_t$ has been substituted for a randomly sampled word from the
vocabulary: $\mathcal{S}^\prime = \left ( w_1, \dots, w_c, \dots, w_n
\,|\, w_c \sim \mathcal{V} \right )$. In this way, every word $w$ is
more predictive of its local context than any other random word in the
corpus.

Every word in $\mathcal{V}$ is mapped to a real-valued vector in
$\Omega$ via a mapping function $C(\cdot)$ such that
$C(w_i) = \left < \mathbf{M}_{\star i} \right >$, where
$\mathbf{M} \in \mathbb{R}^{D \times |\mathcal{V}|}$ is the embedding
matrix and $\left < \mathbf{M}_{\star i} \right >$ is the $i$th column of
$\mathbf{M}$. The network takes $\mathcal{S}$ as input by
concatenating the vectors of the words found in it;
$\mathbf{s}_t = \left < C(w_1)^{\intercal} \| \dots \|
  C(w_t)^{\intercal} \| \dots \| C(w_n)^{\intercal} \right > \in
\mathbb{R}^{nD}$. Similarly, $\mathcal{S}^\prime$ is formed by
substituting $C(w_t)$ for
$C(w_c) \sim \mathbf{M} \, | \, w_c \neq w_t$.

The input vector is then passed through a hard~$\tanh$ layer defined
as,

\begin{align}
  \label{eq:1}
  htanh (x) = \begin{cases} 
    -1 & x < -1 \\
    x & -1 \leqslant x \leqslant 1 \\
    1 & x > 1
  \end{cases} 
\end{align}

\noindent which feeds a single linear unit in the output layer. The
function that is computed by the network is ultimately given by
\eqref{eq:cw_net}:

\begin{align}
  \label{eq:input_activation} 
  \mathbf{s}_t &= \left < \mathbf{M}_{\star 1}^{\intercal} \| \dots \|
                 \mathbf{M}_{\star t}^{\intercal} \| \dots \| \mathbf{M}_{\star n}^{\intercal}
                 \right >^{\intercal} \\
  \mathbf{i} &= \sigma(\mathbf{W}_{hi}\mathbf{s}_t + \mathbf{b}_h) \\
  \label{eq:cw_net}
  f(\mathbf{s}_t) &= \mathbf{W}_{oh} \mathbf{i} + \mathbf{b}_o 
\end{align}

\begin{align*}
  f(\mathbf{s}), \mathbf{b}_o &\in \mathbb{R}^1 \\
  \mathbf{W}_{oh} &\in \mathbb{R}^{H\times 1} \\
  \mathbf{W}_{hi} &\in \mathbb{R}^{D\times H} \\
  \mathbf{s} & \in \mathbb{R}^D \\
  \mathbf{b}_o & \in \mathbb{R}^H \\
\end{align*}

\noindent where $\mathbf{M}, \mathbf{W}_{oh}, \mathbf{W}_{hi},
\mathbf{b}_o, \mathbf{b}_h$ are learnable parameters, $D, H$ are
hyperparameters controlling the size of the input and the hidden
layer, respectively; $\sigma$ is the application of an element-wise
non-linear function ($htanh$ in this case).

The model learns word embeddings by ranking the activation of the true
sequence $\mathcal{S}$ higher than the activation of its `noisy'
counterpart $\mathcal{S}^{\prime}$. The objective of the model then
becomes to minimize the hinge loss which ensures that the activations
of the original and `noisy' $n$grams will differ by at~least~1:

\begin{align}
  \label{eq:error_ngrams}
  \begin{split}
  loss&_{context}(target,\,\,corrupt) = \\ &[1 - f(\mathbf{s}_t) +
  f(\mathbf{s}_{ck}) ]_+, \,\, \forall k \in \mathbb{Z}^E
  \end{split}
\end{align}
\noindent where $E$ is another hyperparameter controlling the number
of `noisy' sequences we give along with the correct sequence
\cite{Mikolov:2013uz,Gutmann:2012:NEU:2188385.2188396}.

\subsection{Augmented C\&W model}
\label{sec:augmented-cw-model}

Following \newcite{Tang_2015}, we extend the previous model to capture
not only the local linguistic environment of each word, but also how
each word contributes to the overall score of the essay. The aim here
is to construct representations which, along with the linguistic
information given by the linear order of the words in each sentence,
are able to capture \textit{usage} information. Words such as
\textit{is, are, to, at} which appear with any essay score are
considered to be \textit{under-}informative in the sense that they
will activate equally both on high and low scoring essays. Informative
words, on the other hand, are the ones which would have an impact on
the essay score (e.g., spelling mistakes).

In order to capture those \textit{score-specific word embeddings}
(\texttt{SSWE}s), we extend (\ref{eq:cw_net}) by adding a further
linear unit in the output layer that performs linear regression,
predicting the essay score. Using (\ref{eq:input_activation}), the
activations of the network (presented in Fig.~\ref{fig:sswe}) are
given by:
\begin{align}
  f_{ss}(\mathbf{s}) &= \mathbf{W}_{oh_1} \mathbf{i} + \mathbf{b}_{o_1} \\
  f_{context}(\mathbf{s}) &= \mathbf{W}_{oh_2} \mathbf{i} +
                            \mathbf{b}_{o_2}
\end{align}
\begin{align*}
  f_{ss}(\mathbf{s}) &\in \left [ \min (score),\, \max (score)  \right
                       ] \\
  \mathbf{b}_{o_1} &\in \mathbb{R}^1 \\
  \mathbf{W}_{oh_1} &\in \mathbb{R}^{1 \times H}
\end{align*}
The error we minimize for $f_{ss}$ (where $ss$ stands for
\textit{score specific}) is the \textit{mean squared error} between
the predicted $\hat{y}$ and the actual essay score $y$:

\begin{align}
  \label{eq:error_scores}
  loss_{score}(\mathbf{s}) = \frac{1}{N} \sum_{i=1}^{N} (\hat{y}_{i}
  - y_i)^2
\end{align}

\begin{figure*}[t]
  \centering
  \subfloat[][Standard neural embeddings]{%
    \begin{tikzpicture}[line cap=round,line join=round,>=triangle 45,x=1.0cm,y=1.0cm,scale=0.9, every node/.style={scale=0.9}]
\draw[color=black] (0.,0.) -- (5.,0.);
\foreach \x in {,1,2,3,4}
\draw[shift={(\x,0)},color=black] (0pt,2pt) -- (0pt,-2pt) node[below] {\footnotesize $\x$};
\draw[color=black] (0.,0.) -- (0.,4.);
\foreach \y in {,1,2,3}
\draw[shift={(0,\y)},color=black] (2pt,0pt) -- (-2pt,0pt) node[left] {\footnotesize $\y$};
\clip(0.,0.) rectangle (5.,4.);
\begin{scriptsize}
\draw [fill=qqqqff] (2.924473300119509,2.291199562552725) circle (2.5pt);
\draw[color=black] (3.7560238770519665,2.291199562552725) node {COPMUTAR};
\draw [fill=qqqqff] (2.341372146254595,2.4994499746473378) circle (2.5pt);
\draw[color=black] (2.6329227231870522,2.686875345532489) node {COMPUTER};
\draw [fill=qqqqff] (1.5812581421092604,1.5623231202215817) circle (2.5pt);
\draw[color=black] (1.810333595413334,1.749748491106733) node {LAPTOP};
\draw [fill=qqqqff] (2.0289965281126765,1.2082974196607406) circle (2.5pt);
\draw[color=black] (2.25807198141675,1.3957227905458915) node {LABTOP};
\end{scriptsize}
\end{tikzpicture}%
    \label{fig:standard_embeddings}}\quad\quad
  \subfloat[][Score-specific word embeddings]{%
    \begin{tikzpicture}[line cap=round,line join=round,>=triangle
  45,x=1.0cm,y=1.0cm, scale=0.9, every node/.style={scale=0.9}]
\draw[color=black] (0.,0.) -- (5.,0.);
\foreach \x in {,1,2,3,4}
\draw[shift={(\x,0)},color=black] (0pt,2pt) -- (0pt,-2pt) node[below] {\footnotesize $\x$};
\draw[color=black] (0.,0.) -- (0.,4.);
\foreach \y in {,1,2,3}
\draw[shift={(0,\y)},color=black] (2pt,0pt) -- (-2pt,0pt) node[left] {\footnotesize $\y$};
\clip(0.,0.) rectangle (5.,4.);
\begin{scriptsize}
\draw [fill=qqqqff] (3.570049577612807,0.5939587039816339) circle (2.5pt);
\draw[color=black] (3.861600154545264,0.7813840748667851) node {COPMUTAR};
\draw [fill=qqqqff] (1.435482853643032,3.478226911492016) circle (2.5pt);
\draw[color=black] (1.7270334305754889,3.665652282377167) node {COMPUTER};
\draw [fill=qqqqff] (1.2168199209436892,3.1346137315359055) circle (2.5pt);
\draw[color=black] (1.8058953742477625,3.1346137315359055) node {LAPTOP};
\draw [fill=qqqqff] (2.924473300119509,0.22952048281606213) circle (2.5pt);
\draw[color=black] (3.5335487534235824,0.22952048281606213) node {LABTOP};
\end{scriptsize}
\end{tikzpicture}%
    \label{fig:ssw_embeddings}}\quad\quad
  \caption{Comparison between standard and score-specific word
    embeddings. By virtue of appearing in similar environments,
    standard neural embeddings will place the correct and the
    incorrect spelling closer in the vector space. However, since the
    mistakes are found in lower scoring essays, \texttt{SSWE}s are
    able to discriminate between the correct and the incorrect
    versions without loss in contextual meaning.}
  \label{fig:embeddings}
\end{figure*}
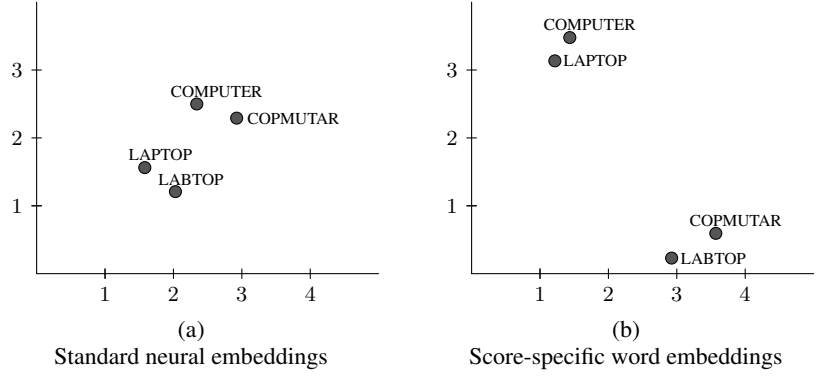

From \eqref{eq:error_ngrams} and \eqref{eq:error_scores} we compute
the overall loss function as a weighted linear combination of the two
loss functions (\ref{eq:loss_function}), back-propagating the error
gradients to the embedding matrix $\mathbf{M}$:

\begin{align}
  \label{eq:loss_function}
  loss_{overall}(\mathbf{s}) &= \begin{aligned}
    \alpha &\cdot loss_{context}(\mathbf{s}, \mathbf{s^\prime}) \\
    &+ (1 - \alpha) \cdot loss_{score} (\mathbf{s})
    \end{aligned}
\end{align}

\noindent where $\alpha$ is the hyper-parameter determining how the
two error functions should be weighted. $\alpha$ values closer to $0$
will place more weight on the score-specific aspect of the embeddings,
whereas values closer to $1$ will favour the contextual information.

Fig.~\ref{fig:embeddings} shows the advantage of using \texttt{SSWE}s
in the present setting. Based solely on the information provided by
the linguistic environment, words such as \textit{computer} and
\textit{laptop} are going to be placed together with their mis-spelled
counterparts \textit{copmuter} and \textit{labtop}
(Fig.~\ref{fig:standard_embeddings}). This, however, does not reflect
the fact that the mis-spelled words tend to appear in lower scoring
essays. Using \texttt{SSWE}s, the correctly spelled words are pulled
apart in the vector space from the incorrectly spelled ones,
retaining, however, the information that \textit{labtop} and
\textit{copmuter} are still contextually related
(Fig.~\ref{fig:ssw_embeddings}).

\subsection{Long-Short Term Memory Network}
\label{sec:lstm}

We use the \texttt{SSWE}s obtained by our model to derive continuous
representations for each essay.  We treat each essay as a sequence of
tokens and explore the use of uni- and bi-directional
\cite{Graves_2012} Long-Short Term Memory networks (LSTMs)
\cite{Hochreiter:1997fq} in order to embed these sequences in a vector
of fixed size. Both uni- and bi-directional LSTMs have been
effectively used for embedding long sequences
\cite{1506.03340v3}. LSTMs are a kind of recurrent neural network
(RNN) architecture in which the output at time $t$ is conditioned on
the input $\mathbf{s}$ both at time $t$ and at time $t-1$:

\begin{align}
  \mathbf{y}_t &= \mathbf{W}_{yh}\mathbf{h}_t +
                 \mathbf{b}_y \label{eq:rnn_y} \\
  \mathbf{h}_t &= \mathcal{H}(\mathbf{W}_{hs}\mathbf{s}_t +
                 \mathbf{W}_{hh}\mathbf{h}_{t-1}+\mathbf{b}_h) \label{eq:rnn_h}
\end{align}

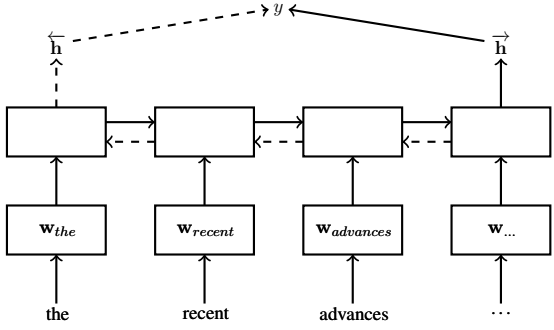
\begin{figure}[t]
  \centering
\begin{tikzpicture}[thick,scale=0.65, every node/.style={scale=0.65}]
\foreach \x/\xtext in {0/the,3/recent,6/advances,9/\dots} \foreach \y in {0,2} {
	\draw (\x, \y) rectangle (\x+2, \y+1);
	\draw[->] (\x+1, \y-1) -- (\x+1, \y);
	\node (y\x) at (\x+1.03,-1.2) {\xtext};
	\node (y\x) at (\x+1.03, 0.5) {$\mathbf{w}_{\xtext}$};
}
\foreach \x in {0,3,6} \draw[->] (\x+2, 2.7) -- (\x+3, 2.7);
\foreach \x in {9,6,3} \draw[dashed,->] (\x, 2.3) -- (\x-1, 2.3);
\draw[->] (10, 3) -- (10,4);
\node (hfor) at (10,4.3) {$\overrightarrow{\mathbf{h}}$};
\draw[dashed, ->] (1, 3) -- (1,4);
\node (hback) at (1,4.3) {$\overleftarrow{\mathbf{h}}$};
\draw[->] (hfor) -- (5.7,5);
\draw[dashed, ->] (hback) -- (5.3,5);
\node (y) at (5.5, 5) {$y$};
\end{tikzpicture}

  \caption{A single-layer \textit{Long Short Term Memory} (LSTM)
    network. The word vectors $\mathbf{w}_i$ enter the input layer one
    at a time. The hidden layer that has been formed at the last
    timestep is used to predict the essay score using linear
    regression. We also explore the use of bi-directional LSTMs
    (dashed arrows). For `deeper' representations, we can stack more
    LSTM layers after the hidden layer shown here.}
  \label{fig:lstm}
\end{figure}

\noindent where $\mathbf{s}_t$ is the input at time $t$, and
$\mathcal{H}$ is usually an element-wise application of a non-linear
function. In LSTMs, $\mathcal{H}$ is substituted for a composite
function defining $\mathbf{h}_t$ as:

\begin{align}
  \mathbf{i}_t &= \begin{aligned} \sigma(\mathbf{W}_{is}\mathbf{s}_t & +
    \mathbf{W}_{ih}\mathbf{h}_{t-1} + \label{eq:lstm_first}\\ &
    \mathbf{W}_{ic}\mathbf{c}_{t-1} + \mathbf{b}_i) \end{aligned} \\
  \mathbf{f}_t &= \begin{aligned} \sigma(\mathbf{W}_{fs}\mathbf{s}_t & +
    \mathbf{W}_{fh}\mathbf{h}_{t-1} + \\ &
    \mathbf{W}_{fc}\mathbf{c}_{t-1} + \mathbf{b}_f) \end{aligned} \\
  \mathbf{c}_t &= \begin{aligned} \mathbf{i}_t \odot
    g(\mathbf{W}_{cs}\mathbf{s}_t & +
    \mathbf{W}_{ch}\mathbf{h}_{t-1} + \mathbf{b}_c) + \\ & \mathbf{f}_t \odot
    \mathbf{c}_{t-1} \end{aligned} 
\end{align}
\begin{align}
  \mathbf{o}_t &= \begin{aligned} \sigma(\mathbf{W}_{os}\mathbf{s}_t & + 
    \mathbf{W}_{oh}\mathbf{h}_{t-1} + \\ &
    \mathbf{W}_{oc}\mathbf{c}_{t} + \mathbf{b}_o) \end{aligned} \\
  \mathbf{h}_t &= \mathbf{o}_t \odot h(\mathbf{c}_t)
\end{align}

\noindent where $g$, $\sigma$ and $h$ are element-wise non-linear
functions such as the \textit{logistic sigmoid} ($\frac{1}{1+e^{-x}}$)
and the \textit{hyperbolic tangent} ($\frac{e^{2z} - 1}{e^{2z} + 1}$);
$\odot$ is the Hadamard product; $\mathbf{W}, \mathbf{b}$ are the
learned weights and biases respectively; and \textit{i}, \textit{f},
\textit{o} and \textit{c} are the input, forget, output gates and the
cell activation vectors respectively.

Training the LSTM in a uni-directional manner (i.e., from left to
right) might leave out important information about the sentence. For
example, our interpretation of a word at some point $t_i$ might be
different once we know the word at $t_{i+5}$. An effective way to get
around this issue has been to train the LSTM in a bidirectional
manner. This requires doing both a forward and a backward pass of the
sequence (i.e., feeding the words from left to right and from right to
left). The hidden layer element in \eqref{eq:rnn_y} can therefore be
re-written as the concatenation of the forward and backward hidden
vectors:

\begin{align}
  \mathbf{y}_t = \mathbf{W}_{yh} \left (
  \begin{array}{c}
    \mathbf{\overleftarrow{\mathbf{h}}}_t^\intercal \\
    \mathbf{\overrightarrow{\mathbf{h}}}_t^\intercal
  \end{array}
  \right ) + \mathbf{b}_y
\label{eq:birnn_y}
\end{align}

We feed the embedding of each \textit{word} found in each essay to the
LSTM one at a time, zero-padding shorter sequences. We form
$D$-dimensional \textit{essay} embeddings by taking the activation of
the LSTM layer at the timestep where the last word of the essay was
presented to the network. In the case of bi-directional LSTMs, the two
independent passes of the essay (from left to right and from right to
left) are concatenated together to predict the essay score. These
essay embeddings are then fed to a linear unit in the output layer
which predicts the essay score (Fig.~\ref{fig:lstm}). We use the
\textit{mean square error} between the predicted and the gold score as
our loss function, and optimize with RMSprop \cite{1502.04390v2},
propagating the errors back to the \textit{word}
embeddings.\footnote{The maximum time for jointly training a
  particular \texttt{SSWE} + LSTM combination took about 55--60 hours
  on an Amazon EC2 {\tt g2.2xlarge} instance (average time was 27--30
  hours).}

\subsection{Other Baselines}
\label{sec:other_baselines}

We train a Support Vector Regression model (see Section
\ref{sec:data}), which is one of the most widely used approaches in
text scoring.  We parse the data using the RASP parser
\cite{Briscoe2006} and extract a number of different features for
assessing the quality of the essays. More specifically, we use
character and part-of-speech unigrams, bigrams and trigrams; word
unigrams, bigrams and trigrams where we replace open-class words with
their POS; and the distribution of common nouns, prepositions, and
coordinators. Additionally, we extract and use as features the rules
from the phrase-structure tree based on the top parse for each
sentence, as well as an estimate of the error rate based on
manually-derived error rules.

$N$grams are weighted using \textit{tf--idf}, while the rest are
count-based and scaled so that all features have approximately the
same order of magnitude. The final input vectors are unit-normalized
to account for varying text-length biases.

Further to the above, we also explore the use of the Distributed
Memory Model of Paragraph Vectors (PV-DM) proposed by
\newcite{1405.4053v2}, as a means to directly obtain essay
embeddings. PV-DM takes as input word vectors which make up $n$gram
sequences and uses those to predict the next word in the sequence. A
feature of PV-DM, however, is that each `paragraph' is assigned a
unique vector which is used in the prediction. This vector, therefore,
acts as a `memory', retaining information from all contexts that have
appeared in this paragraph. Paragraph vectors are then fed to a linear
regression model to obtain essay scores (we refer to this model as
\texttt{doc2vec}).

Additionally, we explore the effect of our score-specific method for
learning word embeddings, when compared against three different kinds
of word embeddings:

\begin{itemize}

\item \texttt{word2vec} embeddings \cite{Mikolov:2013uz} trained on
  our training set (see Section~\ref{sec:data}).

\item Publicly available \texttt{word2vec} embeddings
  \cite{Mikolov:2013uz} pre-trained on the Google News corpus
  (ca. $100$ billion words), which have been very effective in
  capturing solely contextual information.

\item Embeddings that are constructed on the fly by the LSTM, by
  propagating the errors from its hidden layer back to the embedding
  matrix (i.e., we do not provide any pre-trained word
  embeddings).\footnote{Another option would be to use standard C\&W
    embeddings; however, this is equivalent to using \texttt{SSWE}s
    with $\alpha = 1$, which we found to produce low results.}

\end{itemize}

\section{Dataset} 
\label{sec:data}

The Kaggle dataset contains 12.976 essays ranging from 150 to 550
words each, marked by two raters (Cohen's $\kappa = 0.86$). The essays
were written by students ranging from Grade 7 to Grade 10, comprising
eight distinct sets elicited by eight different prompts, each with
distinct marking criteria and score range.\footnote{Five prompts
  employed a holistic scoring rubric, one was scored with a two-trait
  rubric, and two were scored with a multi-trait rubric, but reported
  as a holistic score \cite{Shermis2012}. }  For our experiments, we
use the resolved combined score between the two raters, which is
calculated as the average between the two raters' scores (if the
scores are close), or is determined by a third expert (if the scores
are far apart).  Currently, the state-of-the-art on this dataset has
achieved a Cohen's $\kappa = 0.81$ (using quadratic weights). However,
the test set was released without the gold score annotations,
rendering any comparisons futile, and we are therefore restricted in
splitting the given training set to create a new test set.

The sets where divided as follows: $80\%$ of the entire dataset was
reserved for training/validation, and $20\%$ for testing. $80\%$ of
the training/validation subset was used for actual training, while the
remaining $20\%$ for validation (in absolute terms for the entire
dataset: $64\%$ training, $16\%$ validation, $20\%$ testing). To
facilitate future work, we release the ids of the validation and test
set essays we used in our experiments, in addition to our source code
and various hyperparameter values.\footnote{The code, by-model
  hyperparameter configurations and the IDs of the testing set are
  available at \url{https://github.com/dimalik/ats/.}}

\section{Experiments}
\label{sec:experiments}
\begin{table*}[t]
  \centering
  \begin{threeparttable}
    \begin{tabular}{|l|c|c|c|c|}\hline
    Model                            & Spearman's $\rho$ & Pearson $r$  & RMSE        & Cohen's $\kappa$ \\\hline
    \tt{doc2vec}                     & $0.62$            & $0.63$       & $4.43$      & $0.85$           \\\hline
    SVM                              & $0.78$            & $0.77$       & $8.85$      & $0.75$           \\\hline
    LSTM                             & $0.59$            & $0.60$       & $6.8 $      & $0.54$           \\\hline
    BLSTM                            & $0.7 $            & $0.5 $       & $7.32$      & $0.36$           \\\hline 
    Two-layer LSTM                   & $0.58$            & $0.55$       & $7.16$      & $0.46$           \\\hline
    Two-layer BLSTM                  & $0.68$            & $0.52$       & $7.31$      & $0.48$           \\\hline
    {\tt word2vec}~+~LSTM            & $0.68$            & $0.77$       & $5.39$      & $0.76$           \\\hline
    {\tt word2vec}~+~BLSTM           & $0.75$            & $0.86$       & $4.34$      & $0.85$           \\\hline
    {\tt word2vec}~+~Two-layer LSTM  & $0.76$            & $0.71$       & $6.02$      & $0.69$           \\\hline
    {\tt word2vec}~+~Two-layer BLSTM & $0.78$            & $0.83$       & $4.79$      & $0.82$           \\\hline
    \wpr~+~Two-layer BLSTM           & $0.79$            & $0.91$       & $3.2 $      & $0.92$           \\\hline
    {\tt SSWE}~+~LSTM                & $0.8 $            & $0.94$       & $2.9 $      & $0.94$           \\\hline
    {\tt SSWE}~+~BLSTM               & $0.8 $            & $0.92$       & $3.21$      & $0.95$           \\\hline
    {\tt SSWE}~+~Two-layer LSTM      & $0.82$            & $0.93$       & $3   $      & $0.94$           \\\hline
    {\tt SSWE}~+~Two-layer BLSTM     & $\mbf{0.91}$      & $\mbf{0.96}$ & $\mbf{2.4}$ & $\mbf{0.96}$     \\\hline
   
 \end{tabular}

 \caption{Results of the different models on the Kaggle dataset. All
   resulting vectors were trained using linear regression. We
   optimized the parameters using a separate validation set (see text)
   and report the results on the test set.}
  \label{tab:results}
\end{threeparttable}
\end{table*}

\subsection{Results}
\label{sec:results}

The hyperparameters for our model were as follows: sizes of the layers
$H$, $D$, the learning rate $\eta$, the window size $n$, the number of
`noisy' sequences $E$ and the weighting factor $\alpha$. Also the
hyperparameters of the LSTM were the size of the LSTM layer $D_{LSTM}$
as well as the dropout rate $r$.  Since the search space would be
massive for grid search, the best hyperparameters were determined
using Bayesian Optimization \cite{1206.2944v2}. In this context, the
performance of our models in the validation set is modeled as a sample
from a Gaussian process (GP) by constructing a probabilistic model for
the error function and then exploiting this model to make decisions
about where to next evaluate the function. The hyperparameters for our
baselines were also determined using the same methodology.

All models are trained on our training set (see
Section~\ref{sec:data}), except the one prefixed `{\tt
  word2vec$_{\textrm{pre-trained}}$}' which uses pre-trained
embeddings on the Google News Corpus. We report the Spearman's rank
correlation coefficient $\rho$, Pearson's product-moment correlation
coefficient $r$, and the {root mean square error} (RMSE) between the
predicted scores and the gold standard on our test set, which are
considered more appropriate metrics for evaluating essay scoring
systems \cite{Yannakoudakis_2015}. However, we also report Cohen's
$\kappa$ with quadratic weights, which was the evaluation metric used
in the Kaggle competition.  Performance of the models is shown in
Table~\ref{tab:results}.

In terms of correlation, SVMs produce competitive results
($\rho = 0.78$ and $r = 0.77$), outperforming \texttt{doc2vec}, LSTM
and BLSTM, as well as their deep counterparts. As described above, the
SVM model has rich linguistic knowledge and consists of hand-picked
features which have achieved excellent performance in similar tasks
\cite{DBLP:conf/acl/YannakoudakisBM11}. However, in terms of RMSE, it
is among the lowest performing models ($8.85$), together with `BLSTM'
and `Two-layer BLSTM'.  Deep models in combination with
\texttt{word2vec} (i.e., `\texttt{word2vec}~+~Two-layer LSTM' and
`\texttt{word2vec}~+~Two-layer BLSTM') and SVMs are comparable in
terms of $r$ and $\rho$, though not in terms of RMSE, where the former
produce better results, with RMSE improving by half
($4.79$). \texttt{doc2vec} also produces competitive RMSE results
($4.43$), though correlation is much lower ($\rho = 0.62$ and
$r = 0.63$).

The two BLSTMs trained with \texttt{word2vec} embeddings are among the
most competitive models in terms of correlation and outperform all the
models, except the ones using pre-trained embeddings and
\texttt{SSWE}s.  Increasing the number of hidden layers and/or adding
bi-directionality does not always improve performance, but it clearly
helps in this case and performance improves compared to their
uni-directional counterparts.

Using pre-trained word embeddings improves the results further. More
specifically, we found `{\tt
  word2vec$_{\textrm{pre-trained}}$}~+~Two-layer BLSTM' to be the best
configuration, increasing correlation to 0.79 $\rho$ and 0.91 $r$, and
reducing RMSE to $3.2$. We note however that this is not an entirely
fair comparison as these are trained on a much larger corpus than our
training set (which we use to train our models). Nevertheless, when we
use our \texttt{SSWE}s models we are able to outperform `{\tt
  word2vec$_{\textrm{pre-trained}}$}~+~Two-layer BLSTM', even though
our embeddings are trained on fewer data points. More specifically,
our best model (`{\tt SSWE}~+~Two-layer BLSTM') improves correlation
to $\rho=0.91$ and $r = 0.96$, as well as RMSE to $2.4$, giving a
maximum increase of around $10\%$ in correlation. Given the results of
the pre-trained model, we believe that the performance of our best
\texttt{SSWE} model will further improve should more training data be
given to it.\footnote{Our approach outperforms all the other models in
  terms of Cohen's $\kappa$ too.}

\subsection{Discussion}  
\label{sec:discussion}

Our \texttt{SSWE}~+~LSTM approach having no prior knowledge of the
grammar of the language or the domain of the text, is able to score
the essays in a very human-like way, outperforming other
state-of-the-art systems. Furthermore, while we tuned the models'
hyperparameters on a separate validation set, we did not perform any
further pre-processing of the text other than simple tokenization.

In the essay scoring literature, text length tends to be a strong
predictor of the overall score.  In order to investigate any possible
effects of essay length, we also calculate the correlation between the
gold scores and the length of the essays. We find that the
correlations on the test set are relatively low
($r = 0.3, \rho = 0.44$), and therefore conclude that there are no
such strong effects.

As described above, we used Bayesian Optimization to find optimal
hyperparameter configurations in fewer steps than in regular grid
search. Using this approach, the optimization model showed some clear
preferences for some parameters which were associated with better
scoring models:\footnote{For the best scoring model the
  hyperparameters were as follows:
  $D = 200, H = 100, \eta = 1e-7, n=9, E=200, \alpha=0.1, D_{LSTM} =
  10, r = 0.5$. } the number of `noisy' sequences $E$,
the weighting factor $\alpha$ and the size of the LSTM layer
$D_{LSTM}$. The optimal $\alpha$ value was consistently set to $0.1$,
which shows that our \texttt{SSWE} approach was necessary to capture
the \textit{usage} of the words.  Performance dropped considerably as
$\alpha$ increased (less weight on \texttt{SSWE}s and more on the
contextual aspect).  When using $\alpha = 1$, which is equivalent to
using the basic C\&W model, we found that performance was considerably
lower (e.g., correlation dropped to $\rho = 0.15$.)

The number of `noisy' sequences was set to $200$, which was the
highest possible setting we considered, although this might be related
more to the size of the corpus (see \newcite{Mikolov:2013uz} for a
similar discussion) rather than to our approach. Finally, the optimal
value for $D_{LSTM}$ was $10$ (the lowest value investigated), which
again may be corpus-dependent.

\section{Visualizing the black box}
\label{sec:viz}

\begin{table*}[t]
  \centering
  \begin{tabular}{|c|}\hline
    \dots way to show that \hlc[red3]{Saeng} is a determined
    \hlc[red4]{.} \dots \\\hline
    \dots sometimes I do \hlc[green4]{.} Being \hlc[red3]{patience} is
    being \dots \\\hline
    \dots which leaves the reader \hlc[red4]{satisfied} \dots\\\hline
    \dots is in this picture the \hlc[red1]{cyclist} is riding a dry
    and area which could mean that it is very \\ and the looks to
    \hlc[red2]{be} going down hill there looks to \hlc[red2]{be} a lot
    of turns \hlc[red3]{.} \dots\\\hline
    \dots The only reason im putting this in my own \hlc[red2]{way} is
    because know one is \\ \hlc[red2]{patient} in my family \hlc[green1]{.}
    \dots \\\hline
    \dots \hlc[green1]{Whether} \hlc[green4]{they} are building
    hand-eye coordination \hlc[red1]{,} \hlc[green4]{researching} a
    country \hlc[red1]{,} or \hlc[green1]{family} \hlc[red1]{and}\\
    \hlc[red1]{friends} through @CAPS3 , @CAPS2 , @CAPS6 the
    \hlc[red1]{internet} is \hlc[red3]{highly} \hlc[red1]{and}\\
    \hlc[green2]{I} hope you feel the same way \hlc[green2]{.}\\\hline
  \end{tabular}
  \caption{Several example visualizations created by our LSTM. The
    full text of the essay is shown in black and the `quality' of the
    word vectors appears in color on a range from dark red (low
    quality) to dark green (high quality).}
  \label{tab:viz}
\end{table*}

In this section, inspired by recent advances in
\mbox{(de-)}~convolutional neural networks in computer vision
\cite{1312.6034v2} and text summarization \cite{1406.3830v1}, we
introduce a novel method of generating interpretable visualizations of
the network's performance. In the present context, this is
particularly important as one advantage of the manual methods
discussed in \S~\ref{sec:related_work} is that we are able to know on
what grounds the model made its decisions and which features are most
discriminative.

At the outset, our goal is to assess the `quality' of our word
vectors. By `quality' we mean the level to which a word appearing in a
particular context would prove to be problematic for the network's
prediction. In order to identify `high' and `low' quality vectors, we
perform a single pass of an essay from left to right and let the LSTM
make its score prediction. Normally, we would provide the gold scores
and adjust the network weights based on the error gradients. Instead,
we provide the network with a \textit{pseudo-score} by taking the
maximum score this specific essay can take\footnote{Note the in the
  Kaggle dataset essays from different essay sets have different
  maximum scores. Here we take as $\tilde{y}_{\max}$ the essay set
  maximum rather than the global maximum.} and provide this as the
`gold' score. If the word vector is of `high' quality (i.e.,
associated with higher scoring texts), then there is going to be
little adjustment to the weights in order to predict the highest score
possible. Conversely, providing the minimum possible score (here $0$),
we can assess how `bad' our word vectors are. Vectors which require
minimal adjustment to reach the lowest score are considered of `lower'
quality. Note that since we do a complete pass over the network
(without doing any weight updates), the vector quality is going to be
\textit{essay dependent}.

Concretely, using the network function $f(\mathbf{x})$ as computed by
Eq.~(\ref{eq:lstm_first})~--~(\ref{eq:birnn_y}), we can approximate
the loss induced by feeding the \textit{pseudo-scores} by taking the
\textit{magnitude} of each error vector
(\ref{eq:approxloss1})~--~(\ref{eq:approxloss2}). Since
$\lim_{\|\mathbf{w}\|_2 \rightarrow 0} \hat{y} = y$, this magnitude
should tell us how much an embedding needs to change in order to
achieve the gold score (here pseudo-score). In the case where we
provide the minimum as a \textit{pseudo-score}, a $\|\mathbf{w}\|_2$
value closer to zero would indicate an incorrectly used word. For the
results reported here, we combine the magnitudes produced from giving
the maximum and minimum \textit{pseudo-scores} into a single score,
computed as
$L(\tilde{y}_{\max}, f(\mathbf{x})) - L(\tilde{y}_{\min},
f(\mathbf{x}))$, where:

\begin{align}
  L(\tilde{y}, f(\mathbf{x})) &\approx
                                \|\mathbf{w}\|_2 \label{eq:approxloss1}
  \\
  \mathbf{w} &= \nabla L(\mathbf{x}) \triangleq \left . \frac{\partial
               L}{\partial \mathbf{x}} \right |_{(\tilde{y},
               f(\mathbf{x}))}
\label{eq:approxloss2}
\end{align}

\noindent where $\|\mathbf{w}\|_2$ is the vector Euclidean norm
$\mathbf{w} = \sqrt{\sum_{i=1}^N w_i^2}$; $L(\cdot)$ is the
\textit{mean squared error} as in Eq.~(\ref{eq:error_scores}); and
$\tilde{y}$ is the essay \textit{pseudo-score}.

We show some examples of this visualization procedure in
Table~\ref{tab:viz}. The model is capable of providing positive
feedback. Correctly placed punctuation or long-distance dependencies
(as in Sentence 6 \textit{are \dots researching}) are particularly favoured
by the model. Conversely, the model does not deal well with proper
names, but is able to cope with POS mistakes (e.g., \textit{Being
  patience} or \textit{the internet is highly and \dots}). However, as
seen in Sentence 3 the model is not perfect and returns a false
negative in the case of \textit{satisfied}.

One potential drawback of this approach is that the gradients are
calculated only after the end of the essay. This means that if a
word appears multiple times within an essay, sometimes correctly and
sometimes incorrectly, the model would not be able to distinguish
between them. Two possible solutions to this problem are to either
provide the gold score at each timestep which results into a very
computationally expensive endeavour, or to feed sentences or phrases
of smaller size for which the scoring would be more
consistent.\footnote{We note that the same visualization technique can
  be used to show the `goodness' of phrases/sentences. Within the
  phrase setting, after feeding the last word of the phrase to the
  network, the LSTM layer will contain the phrase embedding. Then, we
  can assess the `goodness' of this embedding by evaluating the error
  gradients after predicting the highest/lowest score.}

\section{Conclusion}
\label{sec:conclusion}

In this paper, we introduced a deep neural network model capable of
representing both local contextual and usage information as
encapsulated by essay scoring. This model yields
\textit{score-specific word embeddings} used later by a recurrent
neural network in order to form essay representations.

We have shown that this kind of architecture is able to surpass
similar state-of-the-art systems, as well as systems based on manual
feature engineering which have achieved results close to the upper
bound in past work. We also introduced a novel way of exploring the
basis of the network's internal scoring criteria, and showed that such
models are interpretable and can be further exploited to provide
useful feedback to the author.

\section*{Acknowledgments}

The first author is supported by the Onassis Foundation. We would like to
thank the three anonymous reviewers for their valuable feedback.

\bibliographystyle{acl2016}
\bibliography{acl2016}

\end{document}